\documentclass[conference]{IEEEtran}

\usepackage{cite}
\usepackage{amsmath,amssymb,amsfonts}
\usepackage{algorithmic}
\usepackage{graphicx}
\usepackage{textcomp}
\usepackage{xcolor}
\def\BibTeX{{\rm B\kern-.05em{\sc i\kern-.025em b}\kern-.08em
    T\kern-.1667em\lower.7ex\hbox{E}\kern-.125emX}}

\definecolor{wong-black}        {HTML}{000000}
\definecolor{wong-lightorange}  {HTML}{E69F00}
\definecolor{wong-lightblue}    {HTML}{56B4E9}
\definecolor{wong-green}        {HTML}{009E73}
\definecolor{wong-yellow}       {HTML}{F0E442}
\definecolor{wong-darkblue}     {HTML}{0072B2}
\definecolor{wong-darkorange}   {HTML}{D55E00}
\definecolor{wong-pink}         {HTML}{CC79A7}
\usepackage{csquotes}
\usepackage{booktabs}
\usepackage[hyperfootnotes=false]{hyperref}
\hypersetup{
    colorlinks=true,
    citecolor=wong-green,
    linkcolor=wong-darkblue,
    filecolor=wong-pink,      
    urlcolor=wong-black,
}
\usepackage[T1]{fontenc}
    
\begin{document}
\bstctlcite{IEEEexample:BSTcontrol}

\title{MUVO: A Multimodal Generative World Model\\for Autonomous Driving\\with Geometric Representations}

\author{\IEEEauthorblockN{Daniel Bogdoll\IEEEauthorrefmark{2}\IEEEauthorrefmark{3}\textsuperscript{\textasteriskcentered},
    Yitian Yang\IEEEauthorrefmark{2}\IEEEauthorrefmark{3}\textsuperscript{\textasteriskcentered},
    Tim Joseph\IEEEauthorrefmark{2},
    Melih Yazgan\IEEEauthorrefmark{2},
    and J. Marius Zöllner\IEEEauthorrefmark{2}\IEEEauthorrefmark{3}}

  \IEEEauthorblockA{\IEEEauthorrefmark{2}FZI Research Center for Information Technology, Germany\\
    bogdoll@fzi.de}
  \IEEEauthorblockA{\IEEEauthorrefmark{3}Karlsruhe Institute of Technology, Germany}}

\maketitle

\def\thefootnote{\textsuperscript{\textasteriskcentered}}\footnotetext{These authors contributed equally}\def\thefootnote{\arabic{footnote}}

\begin{abstract}
World models for autonomous driving have the potential to dramatically improve the reasoning capabilities of today's systems. However, most works focus on camera data, with only a few that leverage lidar data or combine both to better represent autonomous vehicle sensor setups. In addition, raw sensor predictions are less actionable than 3D occupancy predictions, but there are no works examining the effects of combining both multimodal sensor data and 3D occupancy prediction. In this work, we perform a set of experiments with a MUltimodal World Model with Geometric VOxel representations (MUVO) to evaluate different sensor fusion strategies to better understand the effects on sensor data prediction. We also analyze potential weaknesses of current sensor fusion approaches and examine the benefits of additionally predicting 3D occupancy.
\end{abstract}

\section{Introduction}
\label{sec:intro}

In machine learning, recent world models like Cosmos~\cite{nvidia2025cosmosworldfoundationmodel} or Genie~\cite{bruce2024geniegenerativeinteractiveenvironments,deepmind_genie2} have demonstrated the capability to take sequences of high-resolution input images, conditioned on instructions or actions, and generate future images, illustrating possible future scenarios. In autonomous driving, the majority of such world models focus on camera-based inputs~\cite{zhou2025hermes,ma2024unleashinggeneralizationendtoendautonomous,popov2024mitigatingcovariateshiftimitation,gao2024vista,yang2024genad,wang2024driving,huGAIA1GenerativeWorld2023,tesla2023} and some that work in lidar-space~\cite{zhangLearningUnsupervisedWorld2023,chen2024trendunsupervised3drepresentation,zyrianov2024lidardmgenerativelidarsimulation}. These works neglect typical sensor setups of autonomous vehicles. Only two recent works leverage both camera and lidar data~\cite{zhang2024bevworldmultimodalworldmodel,wu2024holodriveholistic2d3dmultimodal}. However, they rely on Bird's-Eye-View (BEV) features as part of their sensor fusion strategy, which is an acknowledged bottleneck due to missing height information~\cite{zhang2024bevworldmultimodalworldmodel}. Finally, world models that predict future 3D occupancies have been proposed, which are highly actionable but rely on visual inputs only~\cite{yang2024driving} or operate in the occupancy space alone~\cite{wang2024occsora}.

In this work, we choose a simple model architecture to perform an extensive set of experiments with varying sensor fusion strategies that do not rely on BEV features and compare them against multiple BEV-based baselines. This way we want to shine some light on how sensor fusion strategies impact the prediction quality of world models. While the benefit of leveraging both camera and lidar data is well-established~\cite{zhang2024fusionocc, liuBEVFusionMultiTaskMultiSensor2023,chittaTransFuserImitationTransformerBased2022}, previous works did not evaluate the impact on future predictions. Our experiment setup takes camera images and lidar point clouds as inputs to predict future observations conditioned on actions. In addition, we examine the effect of an additional decoder component for the prediction of 3D occupancies. Our contributions are:

\begin{itemize}
    \item A multimodal world model with geometric representations that does not rely on BEV features as a bottleneck
    \item Extensive evaluation of sensor fusion strategies based on the prediction quality of the world model
\end{itemize}

The code and model weights are available on~\href{https://github.com/fzi-forschungszentrum-informatik/muvo}{GitHub}\footnote{\href{https://github.com/fzi-forschungszentrum-informatik/muvo}{github.com/fzi-forschungszentrum-informatik/muvo}}.

\begin{figure}[tbp]
  \centering
  \includegraphics[width=\columnwidth]{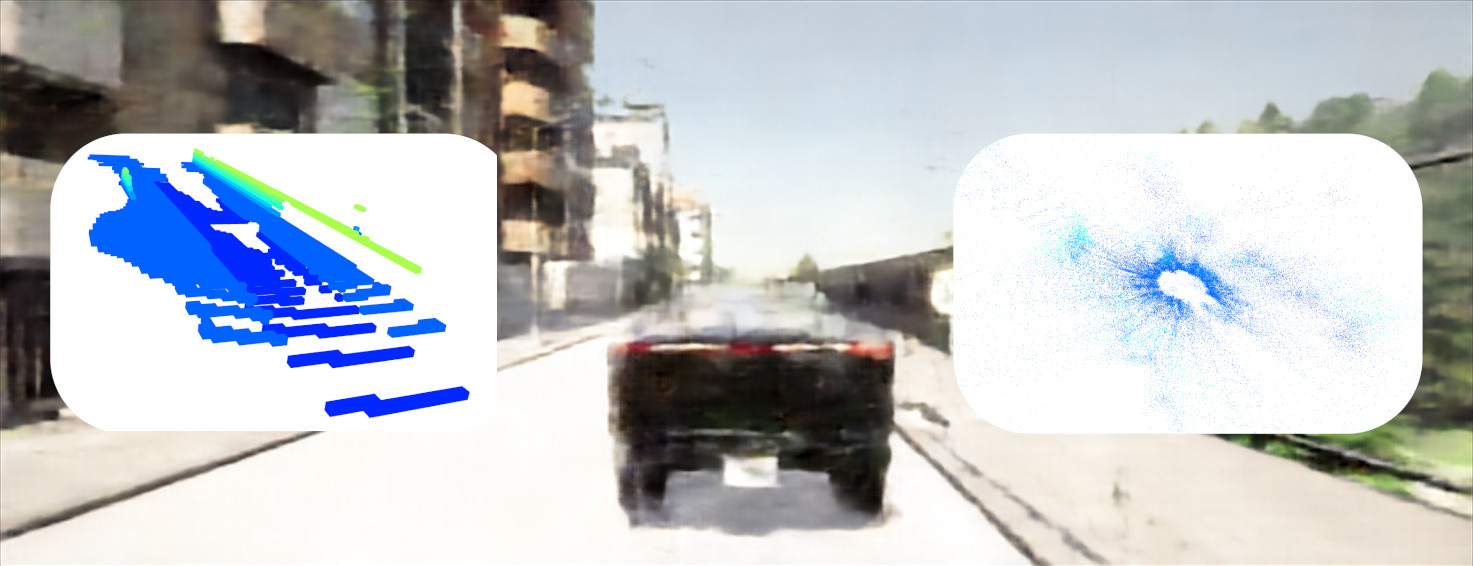}
  \caption{Qualitative output of a sensor fusion experiment with occupancy prediction activated. The predictions shown for camera and lidar sensors and 3D occupancy are based on past camera and lidar inputs.}
  \label{fig:teaser}
\end{figure}
\section{Related Work}
\label{sec:related_work}

\textbf{World Models.} World models are generative models that embed observations into latent states, predict future states conditioned on actions, and decode these latent predictions into the observation space~\cite{lecun, Bogdoll_Exploring_2023_SSCI}.

\begin{figure*}[t!]
  \centering
  \includegraphics[width=1\textwidth]{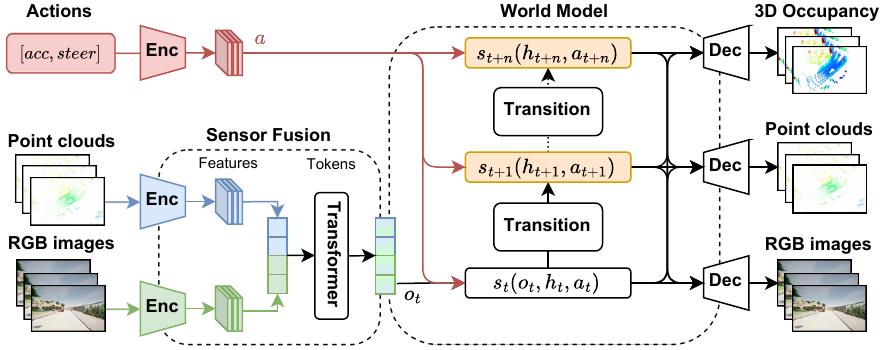}
  \caption{\textbf{MUVO Overview:} Raw camera images and lidar point clouds are processed and fused. The resulting latent representations are fed into our transition model. Conditioned on actions, future states are predicted. Finally, future states are decoded into 3D occupancy grids, raw point clouds, and raw images.}
  \label{fig:overview}
\end{figure*}

Many approaches rely on labels, privileged information, or expert-designed state spaces, limiting their ability to scale. A typical category is the prediction of BEV semantic labels based on supervision during training~\cite{huModelBasedImitationLearning2022,wang2025adawmadaptiveworldmodel,gaoEnhanceSampleEfficiency2022}. DriveDreamer~\cite{wangDriveDreamerRealworlddrivenWorld2023} conditions real-world RGB images on HD maps and labeled 3D bounding boxes. Based on a diffusion model~\cite{rombachHighResolutionImageSynthesis2022}, future frames and actions are jointly predicted. The style of predictions is guided by CLIP~\cite{radfordLearningTransferableVisual2021a} embeddings, using annotated scenes during training. Contrary to these approaches, MUVO does not require labeled training data.

There also exist self-supervised world models. DreamerV3 is capable of predicting futures in Minecraft~\cite{hafnerMasteringDiverseDomains2023}. DriveGAN~\cite{Kim2021_DriveGAN} was trained on real-world data and acts as an action-conditioned neural simulator. DayDreamer~\cite{wu2022daydreamer,hafnerMasteringAtariDiscrete2021} learns robotic tasks from real-world visual inputs. A world model from Tesla~\cite{tesla2023}, trained on proprietary multi-camera RGB data, was demonstrated to predict future observations and semantic or spatial data based on supervised fine-tuning. Following a recent line of work interpreting world models as a single sequence model~\cite{chenDecisionTransformerReinforcement2021,jannerOfflineReinforcementLearning2021,micheliTransformersAreSampleEfficient2022,reedGeneralistAgent2022,liuMaskedAutoencodingScalable2023,wu2023mtm,commaaiCommavqDatasetTokenized2023}, GAIA-1~\cite{huGAIA1GenerativeWorld2023} was trained on proprietary real-world camera data and can be conditioned with both actions and textual inputs. Vector quantization~\cite{vandenoordNeuralDiscreteRepresentation2017} was used to tokenize the data. Based on a video diffusion decoder, it achieved temporally consistent, high-resolution predictions. Similarly, VISTA~\cite{gao2024vista} further increased the image resolution. Compared to recent models of these categories, MUVO is computationally efficient and does not require hundreds of GPUs for training. 

In spatial domains, several world models exist for lidar-data~\cite{zhangLearningUnsupervisedWorld2023,chen2024trendunsupervised3drepresentation,zyrianov2024lidardmgenerativelidarsimulation} or 3D occupancy grids~\cite{yang2024driving, wang2024occsora}. MUVO differs from those approaches by leveraging multimodal data. Most similar to our experiment setup, BEVWorld~\cite{zhang2024bevworldmultimodalworldmodel} and HoloDrive~\cite{wu2024holodriveholistic2d3dmultimodal} leverage both camera and lidar data. BEVWorld proposed a multi-model encoder that generates a unified BEV representation. Upsampled voxel features are used to predict camera and lidar data. Differently, HoloDrive has separate models for image and lidar generation and introduces \textit{2D-to-3D} and \textit{3D-to-2D} structures to improve a joint generation leveraging BEV representations. In both cases, BEV features lack height information and are thus a bottleneck. MUVO does not require BEV features.

\textbf{Scene Completion.} MonoScene~\cite{caoMonoSceneMonocular3D2022a} was the first work to infer 3D semantic voxels from a single 2D camera image, reconstructing visible areas and hallucinating occluded ones. Transformer-based architectures~\cite{li2023voxformer,zhangOccFormerDualpathTransformer2023} have improved by using sparse representations or hybrid encoders.  Symphonies~\cite{jiangSymphonize3DSemantic2023} leverages instance queries to model relations between pixels and voxels that belong to the same instance. OccDepth~\cite{miaoOccDepthDepthAwareMethod2023} uses stereo cameras to derive depth, while many others~\cite{liFBOCC3DOccupancy2023,weiSurroundOccMultiCamera3D2023,huangTriPerspectiveViewVisionBased2023,vuMiLOMultitaskLearning, tesla2022, wangPanoOccUnifiedOccupancy2023} utilize surround vision. Finally, also methods based on lidar point clouds exist~\cite{songSemanticSceneCompletion2017,ristSemanticSceneCompletion2021,yanSparseSingleSweep2020a}. Instead of relying on 3D supervision, recent works~\cite{caoSceneRFSelfSupervisedMonocular2023,panRenderOccVisionCentric3D2023,wimbauerScenesDensityFields2023,panUniOccUnifyingVisionCentric2023} utilize neural rendering to create voxel-based meshes. CLONeR~\cite{carlsonCLONeRCameraLidarFusion2023} uses a single camera and lidar frame to predict occupancy grids based on neural rendering. S4C~\cite{haylerS4CSelfSupervisedSemantic2023} predicts semantic occupancy from a single image, relying on supervision. While these works are related, they focus on data completion rather than world modeling, which can be seen as an additional system component to address occlusions and scene visibility as defined by the sensor setup.

\textbf{Forecasting.} The OpenOcc benchmark~\cite{sima2023_occnet} Was the first to include voxelwise flow information, similar to OpenScene~\cite{openscene2023}. An occupancy network by Tesla was demonstrated to predict motion flow vectors for voxels~\cite{tesla2022}. Khurana et al. combined lidar data with motion sensors to predict future 3D occupancy~\cite{khuranaPointCloudForecasting2023a}. Liu et al. introduced the task of occupancy completion and forecasting ~\cite{liuLiDARbased4DOccupancy2023}, whereas others utilize input images to forecast 3D occupancy~\cite{UniWorld,zhang2024efficientoccupancyworldmodel,zhang2024fusionocc,yang2023vidar}. While these works are related, pure forecasting approaches are not action-conditioned and do not take into account how the planned actions of an agent contribute to future predictions.

\section{World Model}
\label{sec:world_model}

\begin{figure*}[t!]
    \centering
    \includegraphics[width=1\textwidth]{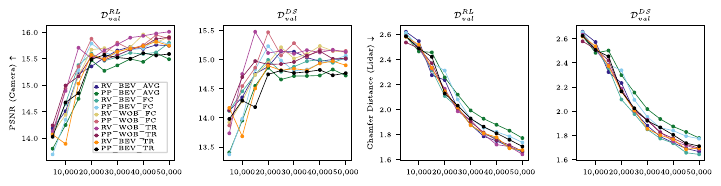}
    \caption{\textbf{Sensor Fusion:} With $\mathcal{{D}}_{{val}}^{{RL}}$ we evaluate representation learning, while $\mathcal{{D}}_{{val}}^{{DS}}$ examines robustness. We examined feature averaging (AVG)~\cite{chenInterpretableEndtoEndUrban2022}, feature concatenation (FC)~\cite{wu2022daydreamer}, and a transformer-based architecture (TR)~\cite{chittaTransFuserImitationTransformerBased2022}. For lidar encodings, we evaluated PointPillars (PP)~\cite{langPointPillarsFastEncoders2019} and a range view (RR)~\cite{liVehicleDetection3D2016} representation followed by a ResNet~\cite{heDeepResidualLearning2016}. For camera, we evaluated direct encoding without BEV (WOB) and a BEV mapping~\cite{philionLiftSplatShoot2020}.}
    \label{fig:sensor_fusion}
\end{figure*}

In this work, we evaluate sensor fusion strategies for world models for autonomous driving. To isolate the effect of fusion strategies, our experiment setup follows the fundamental architecture of MILE~\cite{huModelBasedImitationLearning2022}, which is much reduced in complexity compared to other approaches. As shown in Fig.~\ref{fig:overview}, we introduce changes in the architecture to allow for sensor fusion of a typical sensor setup of autonomous vehicles, comprising stereo cameras and lidar~\cite{Geiger2013IJRR,waymocar}, and predict raw sensor data rather than low-resolution BEV masks based on camera data. In addition, we introduce an additional head to predict 3D occupancy to analyze the effects of introducing a spatial loss in a sensor-independent space. This Multimodal World Model with Geometric Voxel Representations (MUVO) as our setup utilizes image and lidar sensor data from a typical autonomous vehicle setup, as shown in Fig.~\ref{fig:teaser}. First, we process, encode, and fuse RGB camera data and lidar point clouds. Second, we feed the latent representations of the sensor data to a transition model to derive a probabilistic model of the current state, followed by sampling, while concurrently predicting the probabilistic model of future states and sampling from it. Lastly, we decode both current and future states from the probabilistic models, forecasting raw RGB images, point clouds, and 3D occupancy.

\subsection{Observation Encoder}\label{subsec:obs_encoder}
We input RGB images from a front camera and point clouds from a top-mounted lidar. The 3D point cloud, comprising up to $60,000$~points, is projected into a 2D cylindrical projection for lossless bijective representations. For images $\mathcal{I} \in \mathbb{R}^{3\times H_i \times W_i}$, we follow Hu et al~\cite{huModelBasedImitationLearning2022} and use an input size of $600\times960$~pixels.

For images $\mathcal{I}$ and point clouds $\mathcal{R} \in \mathbb{R}^{4 \times H_r \times W_r}$ in range view representation, we utilize a pre-trained backbone for feature extraction. We derive feature maps from different model layers similar to~\cite{huModelBasedImitationLearning2022} and fuse them, culminating in image features $\mathcal{F}_c \in \mathbb{R}^{C \times H_c \times W_c}$ and point cloud features $\mathcal{F}_L \in \mathbb{R}^{C \times H_L \times W_L}$.

\subsection{Multimodal Fusion}\label{subsec: fusion} Similar to~\cite{dosovitskiyImageWorth16x162020,chittaTransFuserImitationTransformerBased2022,shaoSafetyEnhancedAutonomousDriving2022}, we employ the self-attention mechanism of a Transformer~\cite{vaswaniAttentionAllYou2017} to fuse features of different sensors. It takes a sequence of tokens as input, where each token is a $D_t$-dimensional feature vector, so the input sequence is $\mathbf{t}_{in} \in \mathbb{R}^{D_t \times N_t}$, with $D_t$ representing the feature dimension of each token and $N_t$ the number of tokens in the sequence. We flatten the $H$ and $W$ dimensions of the features $\mathcal{F}$ obtained from the encoder described in Sec.~\ref{subsec:obs_encoder}, resulting in tokens $\mathbf{f} \in \mathbb{R}^{C \times HW}$. Subsequently, we incorporate the 2D sinusoidal positional embedding~\cite{vaswaniAttentionAllYou2017,chittaTransFuserImitationTransformerBased2022} $\mathbf{e} \in \mathbb{R}^{C \times HW}$ into each token to introduce spatial inductive biases. The learnable sensor embeddings $\mathbf{s} \in \mathbb{R}^{C \times N_s}$ are added, introducing a sensor category, where $N_s$ is the number of sensors. The resulting tokens $\mathbf{t} \in \mathbb{R}^{C \times HW}$ are obtained, with each token $\mathbf{t}_i(x, y) = \mathbf{f}_i(x, y) + \mathbf{e}_i(x, y) + \mathbf{s}_i$, where $i$ indicates the $i$-th sensor, and $(x, y)$ denotes the coordinate index of that token within the sensor feature.
These tokens from all sensors are concatenated and fed into a Transformer encoder comprising $k$ layers, each consisting of multi-head self-attention, Multilayer Perceptrons~(MLP), and layer normalizations, resulting in new tokens $\mathbf{t}_{new} \in \mathbb{R}^{C \times (\sum_i H_i W_i)}$.

\begin{figure*}[t!]
    \centering
    \includegraphics[width=1\textwidth]{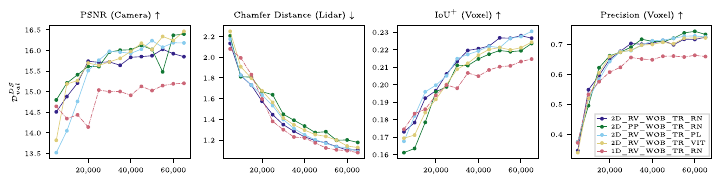}
    \caption{\textbf{Two-dimensional Latent Space}: We compare a 1D baseline to a set of 2D latent spaces, where we also examine the influence of a vision-transformer backbone and an additional perceptual loss term~(\textbf{PL}). For the backbone, we evaluate ResNet18~(\textbf{RN}) and MobileVit-V2~(\textbf{VIT}).}
    \label{fig:latent_space_voxel}
\end{figure*}
    
\subsection{Transition Model}\label{subsec: transition model}
 The input consists of fused observation features $\mathbf{o}_{0:t}$ and encoded actions $\mathbf{a}_{0:t} \in \mathbb{R}^{T \times D_a}$, based on a simple MLP, assuming access to a policy or motion planner. The output includes stochastic hidden states $\mathbf{s}_{0:t} \in \mathbb{R}^{T \times D_s}$ and deterministic historical states $\mathbf{h}_{0:t} \in \mathbb{R}^{T \times D_h}$, predictions for future states $\mathbf{s}_{t:t+n}$, and $\mathbf{h}_{t:t+n}$, $T$ represents the number of frames, also referred to as the sequence length, and $D_a$, $D_s$, $D_h$ are the dimensions of each vector respectively.
    The deterministic historical variable $\mathbf{h}_{t+1} = f_\theta(\mathbf{h}_t, \mathbf{s}_t)$ is modelled by a Gated Recurrent Unit~(GRU)~\cite{choLearningPhraseRepresentations2014} $f_\theta,$ enabling the model to remember past states. The posterior hidden state probability distribution is given by $q(\mathbf{s}_t\mid\mathbf{o}_{\leq t}, \mathbf{a}_{<t}) \sim \mathcal{N}_\phi(\mathbf{o}_t, \mathbf{h}_t, \mathbf{a}_{t})$, while the prior hidden state probability distribution, without the input of observed feature $\mathbf{o}_t$, is given by $p(\mathbf{s}_t\mid\mathbf{h}_t, \mathbf{a}_{t-1}) \sim \mathcal{N}_\theta(\mathbf{h}_t, \mathbf{a}_{t-1})$. Here, $\mathcal{N}_\phi$ and $\mathcal{N}_\theta$ are probability models modelled by a MLP.
    Given observations, $\mathbf{s}_t$ is sampled from the posterior distribution $q$. In the absence of observations, i.e., during prediction, $\hat{\mathbf{s}_t}$ is sampled from the prior distribution $p$.

    We utilize the output tokens $\mathbf{t}_{new}$ from the sensor fusion in the form of a two-dimensional latent state as the encoded observations $\mathbf{o}_t$ for the transition model. Compared to 1D states, we set the stochastic hidden states $\mathbf{s}_t$, and deterministic historical states $\mathbf{h}_t$ to shape $C_h \times (\sum_j T_j)$, where $T_j$ is the number of tokens of each output modality. For $f_\theta$, we replace all fully connected layers with convolutional layers in order to utilize two-dimensional states. The probability models $\mathcal{N}_\phi$ and $\mathcal{N}_\theta$ are modeled by a transformer decoder. Learnable embeddings, which have the same shape as the stochastic hidden states $\mathbf{s}_t$, are used as queries, where $\mathbf{h}_t, \mathbf{a}_{t}, (\mathbf{o}_t)$ are concatenated as key-value pairs. Then, we query the state information in these through the attention mechanism to obtain stochastic hidden state tokens.

\subsection{Multimodal Decoder}\label{subsec: decoder}
    We decode into camera and lidar data and introduce an optional head for 3D occupancy. The input is a latent dynamic state $(\mathbf{s}_t, \mathbf{h}_t)$ with shape $C \times \sum_j{T_j}$ which is provided by the transition model. We divide those tokens based on modalities and reshape each modality's tokens $C \times T_j$ to fit the output shape. The occupancy decoder is of shape $C\times X\times Y \times Z$. For camera and lidar data, first, the input is reshaped to $C \times H_0 \times W_0$, where $H_0$ and $W_0$ are determined by the final output resolution $H \times W$. Subsequently, we perform upscaling with convolutional networks similar to~\cite{hafnerDreamControlLearning2020,ha2018worldmodels} to produce a feature map of size $C_n \times H \times W$. For camera and lidar, we utilize two-dimensional convolutions, while we employ three-dimensional convolutions for voxels.

\section{Evaluation}\label{subsec: evaluation}

\begin{figure*}[t!]
    \centering
    \includegraphics[width=\textwidth]{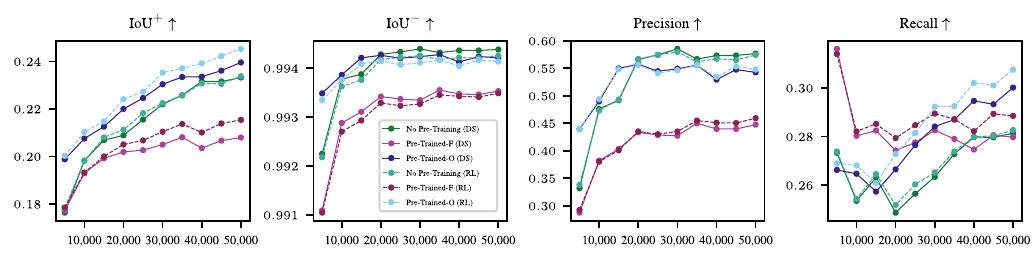}
    \caption{\textbf{Pre-Training:} Influence of camera-lidar pre-training for 50,000 steps on 3D occupancy prediction. We evaluate on both $\mathcal{{D}}_{{val}}^{{RL}}$ and $\mathcal{{D}}_{{val}}^{{DS}}$. The green lines show our benchmark without pre-training. Violet lines show frozen weights of the pre-trained model, and weights remained open for the blue lines.}
    \label{fig:occ_learning}
\end{figure*}

For the evaluation, we first present our utilized training setup in Sec.~\ref{subsec:training_setup}, followed by the evaluation of sensor fusion strategies in Sec.~\ref{subsec: sensor fusion}. Fig.~\ref{fig:sensor_fusion} shows the impact of different fusion strategies, and Fig.~\ref{fig:latent_space_voxel} demonstrates the difference of differently sized latent spaces. Finally, we examine the effects of the optional 3D occupancy prediction in Sec.~\ref{subsec:3d_occupancy}. Fig.~\ref{fig:occ_learning} shows the relation between camera-lidar based pre-training and 3D occupancy prediction, and Fig.~\ref{fig:occ_vox_impact} shows the reverse impact of predicting occupancy on the quality of sensor predictions.

\subsection{Training Setup}\label{subsec:training_setup}

\textbf{Training Losses.} For our experiments, we follow a self-supervised training approach and do not require labels at any point. For each modality, we downsample multiple times with ratios of 1, 2, and 4. With this multi-scale approach, we compute losses at different resolutions. For \textbf{images}, we output RGB data that align with the size of the input and utilize the common L1 loss $\mathcal{L}^\text{img}$ for the minimization of the absolute discrepancies between target and prediction. For \textbf{point clouds}, we generate range view images of dimensions $H_r\times W_r\times 4$, which can be converted into $N\times 3$ point cloud data. The target is the range view image transformed from the ground truth, where an L2 loss $\mathcal{L}^\text{p, xyz}$ is applied to minimize the Euler distance and an L1 loss $\mathcal{L}^\text{p, r}$ is based on range $r$. For \textbf{3D Occupancy,} voxel grids of size $192\times192\times64$ with $0.5m$ voxels contain the binary occupancy. The target is obtained by voxelizing fused depth maps from depth cameras and point clouds from lidar. The loss for voxel grids uses a Scene-Class Affinity Loss~(SCAL)~\cite{caoMonoSceneMonocular3D2022a} $\mathcal{L}^\text{V, scal}$. The total loss is given by
\begin{equation}
\begin{split}
\mathcal{L} = \sum \lambda_i \bigl(
& \lambda_\text{img} \mathcal{L}_i^\text{img} + \lambda_\text{pcd} (\mathcal{L}_i^\text{pxyz} + \mathcal{L}_i^\text{pr} + \mathcal{L}_i^\text{pcd}) + \lambda_\text{V} \mathcal{L}_i^\text{V, scal})
\end{split}
\label{eq:loss}
\end{equation}

\textbf{Datasets.}\label{subsubsec: dataset}
We collected a training dataset $\mathcal{{D}}_{{train}}$ in the CARLA simulation environment~\cite{Dosovitskiy17} using an expert reinforcement learning agent~\cite{zhang2021roach,huModelBasedImitationLearning2022} for a more realistic driving style . Our data collection encompasses four towns (Town01, Town03, Town04, Town06) and four weather conditions (Clear Noon, Wet Noon, Hard Rain Noon, Clear Sunset), gathered at a frequency of 10~FPS. For each town, we executed 25 runs, each lasting 300 seconds, with randomly selected weather conditions, amounting to 300,000 frames of data. We obtain RBG image $\mathcal{I} \in \mathbb{R}^{3 \times 600 \times 960}$, depth map $\mathcal{I}_D \in \mathbb{R}^{1 \times 600 \times 960}$, e.g., derived from stereo cameras, point cloud $\mathcal{P} \in \mathbb{R}^{\leq60,000 \times 3}$ obtained from a lidar with 64 vertical channels, route map $route \in \mathbb{R}^{1 \times 64 \times 64}$ as the planned route in BEV space, speed $\mathbf{v} \in \mathbb{R}$, and actions $\mathbf{a} \in \mathbb{R}^2$ in the form of acceleration and steering angle.

We adopt the same setup as before for two distinct validation sets. For each town, we execute five 300-second long driving sessions with the following settings:
        
        $\mathcal{{D}}_{{val}}^{RL}$: This set uses the same cities and weather conditions as the training set. However, the driving routes are randomized. The goal is to evaluate the effectiveness of our model in representation learning (RL) in familiar environments.
        
        $\mathcal{{D}}_{{val}}^{{DS}}$: We maintain the same cities as in the training set but introduce different weather conditions. The driving routes are also randomized to evaluate the model performance under domain shifts~(DS).
        
    \textbf{Training Parameters.} We sampled data at intervals of 0.2 seconds, creating sequences of length 12 to serve as training inputs. All 12 frames were treated as known data. In the experiments containing voxel reconstructions, we reduced the length of sequences to 6 to speed up the training. We trained with a batch size of 16 and the AdamW optimizer with a learning rate of $10^{-4}$ and a weight decay of $0.01$. For validation, we used 6/4 frames as given observations, while 6/2 served as ground truth. For all experiments, we use a pre-trained ResNet18~\cite{heDeepResidualLearning2016} as our baseline backbone.

\subsection{Sensor Fusion Strategies}\label{subsec: sensor fusion}

Several prior multimodal world models rely on naive fusion approaches~\cite{wu2022daydreamer,shresthaSenseImagineAct2023,gaoEnhanceSampleEfficiency2022}. For our experiments, we compare those to a transformer-based architecture. To evaluate the effect of different sensor fusion strategies, we used metrics based on the evaluated modality: For assessing the quality of \textbf{image} predictions, we use the common Peak Signal-to-Noise Ratio~(PSNR) to assess average differences. We use the Chamfer Distance to evaluate the accuracy of \textbf{point cloud} predictions. For the predictions of \textbf{3D occupancy grids}, we used the metrics Intersection over Union~(IoU), Precision, and Recall. We differentiate between IoU$^+$ for occupied voxels and IoU$^-$ for empty ones.

We first present the examined decoders and fusion methods. Subsequently, we provide an overview and a comparison of all analyzed combinations, as shown in Figure~\ref{fig:sensor_fusion}.

    \textbf{Encoders.}\label{subsubsec: extra encoder} For \textit{image} features $\mathcal{F}_c$, we compare the standard encoder introduced in Sec.~\ref{subsec:obs_encoder} to approaches that map features to BEV space~\cite{philionLiftSplatShoot2020,zhang2024bevworldmultimodalworldmodel,huModelBasedImitationLearning2022,liangBEVFusionSimpleRobust2022}. Here, features are first elevated into a 3D space. Then, these 3D feature voxels are aggregated into the BEV space, leading to image features $\mathcal{F}_b \in \mathbb{R}^{C \times H_b \times W_b}$.  We evaluate both lossless and lossy representations for point clouds. We compare a range view-based representation with PointPillars~\cite{langPointPillarsFastEncoders2019} as an encoder, where point clouds are segmented into discrete pillars along the X and Y axes followed by data processing and feature extraction, resulting in a 2D BEV pseudo-image.

        \textbf{Latent Space.}
        \label{subsec:latent}
        In prior works, the latent space was commonly modeled as one-dimensional vectors~\cite{huModelBasedImitationLearning2022,hafnerDreamControlLearning2020}, which may limit model performance by introducing a representational bottleneck. We perform experiments with both a 1D and a 2D latent space. In addition, we examine an additional perceptual loss~\cite{johnson2016perceptual} and a vision transformer backbone~\cite{mehta2022separable}.
        
Figure~\ref{fig:latent_space_voxel} shows four evaluation graphs. It is divided into the prediction performance for camera (a), lidar (b), and 3D occupancy (c,d) on $\mathcal{{D}}_{{val}}^{{DS}}$. The 2D latent state significantly benefits predictions for camera images and spatial voxel occupancies, while lidar predictions do not see any benefit. This might be due to the fact that camera data is much more complex than lidar data. Compared to the baseline 2D model (dark blue), we do not see a strong effect of utilizing a perceptual loss, as it produces visually poorer reconstructions and does not show any significant advantages. Using the vision transformer as an encoder does provide advantages for the prediction of camera images but shows no effect on other metrics. This shows that the 2D latent space itself provides the largest boost in performance, while other changes have little effect.
    
    \textbf{Fusion Methods.} We compare a transformer-based sensor fusion approach, as described in Sec.~\ref{subsec: fusion}, with naive combinations of encoded 1D features from each sensor modality, as found in the literature. We perform experiments for both averaging features as well as concatenating them, followed by a fully connected layer. To generate such latent states, the output tokens are reshaped into their original shape after the encoding, namely $\mathcal{F}_c^{new} \in \mathbb{R}^{C \times H_c \times W_c}$ and $\mathcal{F}_L^{new} \in \mathbb{R}^{C \times H_L \times W_L}$. Each feature is then downsampled by convolutional layers, followed by pooling layers to get one-dimensional features $\mathbf{f}_d \in \mathbb{R}^D$, which are subsequently concatenated and then passed through fully connected layers to reduce its dimensionality, producing the vector $\mathbf{o}_t \in \mathbb{R}^D$.\\

    We evaluate the prediction performance of eight encoder-fusion combinations, as visible in Figure~\ref{fig:sensor_fusion}. We follow a naming scheme A-B-C: \textbf{A} represents the method of processing point clouds: \textit{PP} stands for the use of PointPillars as the encoder; \textit{RV} indicates the conversion of point clouds into range view. \textbf{B} denotes the approach of image processing: \textit{BEV} implies mapping to BEV followed by feature extraction with a backbone; \textit{WOB} denotes that no BEV mapping is performed. \textbf{C} describes the method of sensor fusion: \textit{AVG} stands for the averaging of 1D features; \textit{FC} means that concatenation followed by a fully connected layer is performed; \textit{TR} denotes that the transformer-based multi-head self-attention mechanism was used, as described in Sec.~\ref{subsec: fusion}. In the following, we first discuss the effects on image predictions, followed by the effects on point cloud predictions.


        \textbf{Image Prediction}. The impact of the different experiments on the quality of camera predictions is shown in Fig.~\ref{fig:sensor_fusion}~a) and \ref{fig:sensor_fusion}~b). We observe a drop in performance for all networks in the $\mathcal{{D}}_{{val}}^{{DS}}$ dataset compared to $\mathcal{{D}}_{{val}}^{{RL}}$, but the relative performance of different networks remains consistent across both datasets. Generally, the transformer-based architecture \textit{RV-WOB-TR} performs on par or better compared to the other combinations, and range view-based lidar encodings show clear advantages over PointPillars. Methods with an additional BEV mapping of image features perform worse, and combinations with PointPillars suffer especially. We can see that the effectiveness of introducing a transformer-based architecture depends on the encoder used. It outperforms other approaches when combined with a ResNet-18 for feature extraction. In contrast, when combined with PP and BEV, its performance is lower than  concatenating (FC) but higher than averaging (AVG).

        \textbf{Point Cloud Prediction}. The impact of the different experiments on the quality of camera predictions is shown in Fig.~\ref{fig:sensor_fusion}~c) and \ref{fig:sensor_fusion}~d). Examining the Chamfer Distance plots, where lower values mean better performance, we find no significant performance disparity between both validation datasets. For $\mathcal{{D}}_{{val}}^{{RL}}$, the transformer-based architecture \textit{RV-WOB-TR} performs on par or better compared to the other combinations. However, on $\mathcal{{D}}_{{val}}^{{DS}}$, its performance drops. As before, range view-based methods demonstrate superiority over PointPillars. Utilizing BEV features shows no clear disadvantage for this task. We can see that transformer-based architectures generally outperform other fusion techniques.

        We determine a transformer-based architecture with a 2D latent space and lossless range-view representations for point clouds as an optimal fusion strategy, while performance benefits are more pronounced for camera predictions.

\begin{figure}[t!]
    \centering
    \includegraphics[width=1\columnwidth]{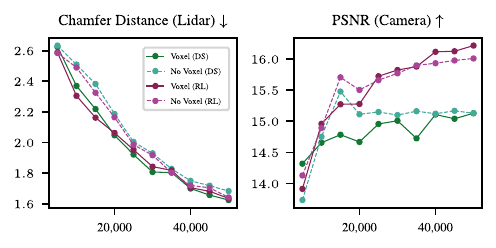}
    
    \caption{\textbf{Occupancy:} Impact of predicting 3D occupancy on the quality of camera and lidar predictions, evaluated on both $\mathcal{{D}}_{{val}}^{{RL}}$ and $\mathcal{{D}}_{{val}}^{{DS}}$.}
    \label{fig:occ_vox_impact}
\end{figure}

\subsection{3D Occupancy Prediction}
\label{subsec:3d_occupancy}
    Next to analyzing fusion strategies, we are interested in the effects of also predicting more actionable 3D occupancies. Our experiments analyze whether an occupancy model can benefit from a pre-trained model which was trained by only predicting camera and lidar data, as shown in Fig.~\ref{fig:occ_learning}.

    Subsequently, we analyze if occupancy prediction improves the prediction of camera and lidar data, as shown in Fig.~\ref{fig:occ_vox_impact}.

    \textbf{3D Occupancy Prediction.} We perform experiments in three scenarios, as shown in Figure~\ref{fig:occ_learning}. As we want to examine the effect of encoded knowledge of predicting camera and lidar data on 3D occupancy, we first train a model as a pre-trained starting point that predicts camera and lidar data alone for 50,000 steps. For the first scenario, we employ the pre-trained model but freeze \textbf{(PTF)} all of its weights so that only the weights of the voxel decoder are trained. This approach allows us to assess the impact of fine-tuning only the voxel-specific aspects of the model while keeping the rest of the network, in particular all encoders, constant to evaluate if any information about a discrete geometry of the world is already encoded based on camera and lidar data. For the second scenario, the pre-trained weights were used as a starting point, but the entire network was open \textbf{(PTO)} for weight updates during training. Here, we analyze how the pre-trained weights influence the learning process when the whole network adapts and evolves during training. For the third scenario, no pre-training \textbf{(NPT)} is utilized, and we train the network from scratch.

    In Figure~\ref{fig:occ_learning} we observe that the model trained from scratch (NPT) exhibits a similar performance on both validation datasets across all four metrics, while the other two models using pre-trained weights (PTF and PTO) generally performed better on $\mathcal{{D}}_{{val}}^{{RL}}$ than on $\mathcal{{D}}_{{val}}^{{DS}}$ across three metrics, excluding IoU$^-$. Interestingly, for IoU$^-$ we observe an opposite behavior, where the models perform better on $\mathcal{{D}}_{{val}}^{{DS}}$. This is attributed to voxel occupancy grid predictions focusing more on occupied grids. Since voxel grids are mostly empty, models on $\mathcal{{D}}_{{val}}^{{DS}}$ tend to predict more noise, leading to lower IoU$^-$ scores.  Comparing PTO to NPT, the PTO model showed advantages early on, supporting the idea that pre-trained weights contribute valuable spatial knowledge. However, in the later stages of training, the NPT model overtook the PTO model in Precision, while the PTO model remained superior for IoU$^-$ and Recall. This indicates that the non-pre-trained model adopts a more conservative strategy for 3D occupancy prediction. When we examine the PTF scenario, although the model underperformed compared to the other two, its performance improved over time by only training the voxel decoder. This improvement underscores that the pre-trained weights already contain some, however limited, spatial information, indicating that the model partially integrates image and point cloud features to form spatial voxel features even when trained only on these two modalities.

    As learning 3D occupancy is computationally intensive, we conclude that pre-training strategies on only camera and lidar data are generally recommendable, as they both speed up training and show overall superior performance.

    \textbf{Sensor Data Predictions.}
    We perform experiments to determine whether knowledge encoded through occupancy can be leveraged by lidar and camera predictions, as shown in Figure~\ref{fig:occ_vox_impact}. Based on the chamfer Distance for point clouds and the PSNR metric for images, we observe only slightly increased performance gains for both modalities when occupancy prediction is included, with a more pronounced benefit for camera predictions under the $\mathcal{{D}}_{{val}}^{{RL}}$ setting.

\section{Conclusion}
\label{sec:conclusion}

We have presented an extensive set of experiments on sensor fusion strategies for predictive world models in autonomous driving. In addition, we have examined the effects of additionally predicting 3D occupancy. This way, the effects of using typical sensor setups combined with learning actionable occupancy data can be better understood.

We were particularly interested in the question of whether the introduction of a BEV feature representation, as typical in the literature~\cite{zhang2024bevworldmultimodalworldmodel,wu2024holodriveholistic2d3dmultimodal}, might act as a bottleneck, as it misses height information. Our experiments demonstrate that lossless lidar representations with a standard transformer-based fusion and an increased 2D latent space are indeed beneficial in the case of camera-lidar fusion. 

When analyzing the effects of introducing actionable but computationally intensive spatial 3D occupancy predictions, our experiments show that occupancy predictions benefit from more efficient pre-training with only camera and lidar predictions. In addition, we observed that occupancy prediction leads to minor improvements for both camera and lidar predictions.

For future work, we would be interested in performing large-scale experiments with real-world data as our approach does not require labels, leading to a better understanding of the implications of utilizing simulated data only. This might include a more complex sensor setup, raising the need to examine computational efficiency more thoroughly.
\section*{Acknowledgment}
\label{sec:acknowledgment}
This work results from the just better DATA project supported by the German Federal Ministry for Economic Affairs and Climate Action (BMWK), grant number 19A23003H.

\bibliographystyle{IEEEtran}
\bibliography{references}

\end{document}